\documentclass[conference]{IEEEtran}
\IEEEoverridecommandlockouts
\usepackage{cite}
\usepackage{amsmath,amssymb,amsfonts}
\usepackage{algpseudocode}
\usepackage{algorithm}
\usepackage{graphicx}
\usepackage{textcomp}
\usepackage{xcolor}
\def\BibTeX{{\rm B\kern-.05em{\sc i\kern-.025em b}\kern-.08em
    T\kern-.1667em\lower.7ex\hbox{E}\kern-.125emX}}
\begin{document}

\title{Decentralized Semantic Traffic Control in AVs Using RL and DQN for Dynamic Roadblocks}


	\author{
	 \IEEEauthorblockN{ Emanuel Figetakis\IEEEauthorrefmark{1}, Yahuza Bello\IEEEauthorrefmark{1}, Ahmed Refaey \IEEEauthorrefmark{1} \IEEEauthorrefmark{2}, and Abdallah Shami\IEEEauthorrefmark{2}} \\

	\IEEEauthorblockA{\IEEEauthorrefmark{1} University of Guelph, Guelph, Ontario, Canada.}
	\IEEEauthorblockA{\IEEEauthorrefmark{2} Western University, London, Ontario, Canada.}}
\maketitle

\begin{abstract}
Autonomous Vehicles (AVs), furnished with sensors capable of capturing essential vehicle dynamics such as speed, acceleration, and precise location, possess the capacity to execute intelligent maneuvers, including lane changes, in anticipation of approaching roadblocks. Nevertheless, the sheer volume of sensory data and the processing necessary to derive informed decisions can often overwhelm the vehicles, rendering them unable to handle the task independently. Consequently, a common approach in traffic scenarios involves transmitting the data to servers for processing, a practice that introduces challenges, particularly in situations demanding real-time processing. In response to this challenge, we present a novel DL-based semantic traffic control system that entrusts semantic encoding responsibilities to the vehicles themselves. This system processes driving decisions obtained from a Reinforcement Learning (RL) agent, streamlining the decision-making process. Specifically, our framework envisions scenarios where abrupt roadblocks materialize due to factors such as road maintenance, accidents, or vehicle repairs, necessitating vehicles to make determinations concerning lane-keeping or lane-changing actions to navigate past these obstacles. To formulate this scenario mathematically, we employ a Markov Decision Process (MDP) and harness the Deep Q Learning (DQN) algorithm to unearth viable solutions.
\end{abstract}


\section{Introduction}

In recent decades, substantial research endeavors have been dedicated to enhancing Intelligent Transportation Systems (ITS). Researchers have introduced a plethora of solutions aimed at enhancing security and comfort in autonomous driving, as documented in several studies \cite{li2023automated,liu2023real}. Nevertheless, a significant portion of these proposed solutions necessitates the exchange of substantial volumes of sensor-derived data among various entities, including vehicles, edge servers, and Road Side Units (RSUs), utilizing diverse communication techniques such as Vehicle-2-Vehicle (V2V) and Vehicle-2-Infrastructure (V2I) methods. Compounding this challenge, the constrained computing resources available in vehicles constrain their ability to perform resource-intensive tasks. 

Semantic communication emerges as a promising candidate for addressing these challenges, particularly in terms of resource limitations and the need to transmit extensive data. This potential stems from its ability to transmit data efficiently even when resources are limited \cite{luo2022semantic}. As a result, semantic communication finds extensive application across various domains, including but not limited to ITS, eXtended Reality (XR), Augmented Reality (AR), Natural Language Processing (NLP), and Industrial Internet of Things (IIoT).

Several semantic communication systems incorporating Deep Learning (DL) have been introduced \cite{xie2021deep,huang2021deep}. Within a DL-driven semantic communication system, semantic information is typically characterized as concealed features residing within observable data. Traditional methods often struggle to extract these features effectively, leading to the adoption of DL techniques, particularly Neural Networks (NN), to estimate and capture these hidden attributes. 
This opens the door to the incorporation of semantic-based communication into numerous practical applications.


In the realm of autonomous driving, a DL-based semantic communication framework holds the potential to be a potent tool for efficiently mitigating the substantial exchange of vast data volumes. Typically, within the literature, researchers often assume that the vehicles in question are equipped with sensors tasked with gathering essential information to facilitate decision-making processes, such as traffic management, lane-changing, and lane-keeping \cite{figetakis2023implicit, wang2021harmonious}. Conventionally, the acquired data is transmitted through conventional communication systems to either a Base Station (BS) or an edge server for the resource-intensive tasks involved in making informed decisions in various driving scenarios. 

Semantic-based communication offers a promising solution to address this challenge by transmitting solely the requisite semantic information gleaned from the driving environment. In this manner, vehicles become capable of handling the semantic encoding aspect of data transmission, thereby enhancing the overall driving experience. Therefore, this paper introduces a DL-based semantic traffic control system that delegates semantic encoding tasks to the vehicles, processing driving decisions acquired from a Reinforcement Learning (RL) agent. 
The contributions of this study are succinctly outlined below.
\begin{itemize}
    \item Propose a DL-based semantic traffic control system that ensures the semantic encoding aspect of the framework is executed within the vehicles to acquire the correct decision made by an RL agent. Specifically, our framework envisions scenarios where sudden roadblocks emerge due to factors such as road maintenance, accidents, or vehicle repairs, requiring vehicles to make decisions regarding lane-keeping or lane-changing actions to navigate past the roadblock.
    \item Formulate an MDP model to depict a hypothetical situation in which vehicles must make suitable decisions when faced with a roadblock on a highway. Assess this model's performance using the widely recognized Deep Q Learning (DQN) algorithm to determine the optimal driving choices. 
\end{itemize}


The remaining sections of the paper are organized as follows: In Section \ref{RW}, we delve into the background and explore related works concerning semantic communication and its application in ITS domain. Section \ref{SM} is devoted to the system model, encompassing the comprehensive model description, MDP formulation elucidation, and the presentation of the RL-based solution approach. In Section \ref{ER}, we provide insights into the experimentation process and analyze the results obtained. Finally, Section \ref{C} serves as the conclusion of the paper, summarizing the key findings and contributions. 


\begin{figure*}[ht]
\centering
   \includegraphics[width=\textwidth, height=7cm]{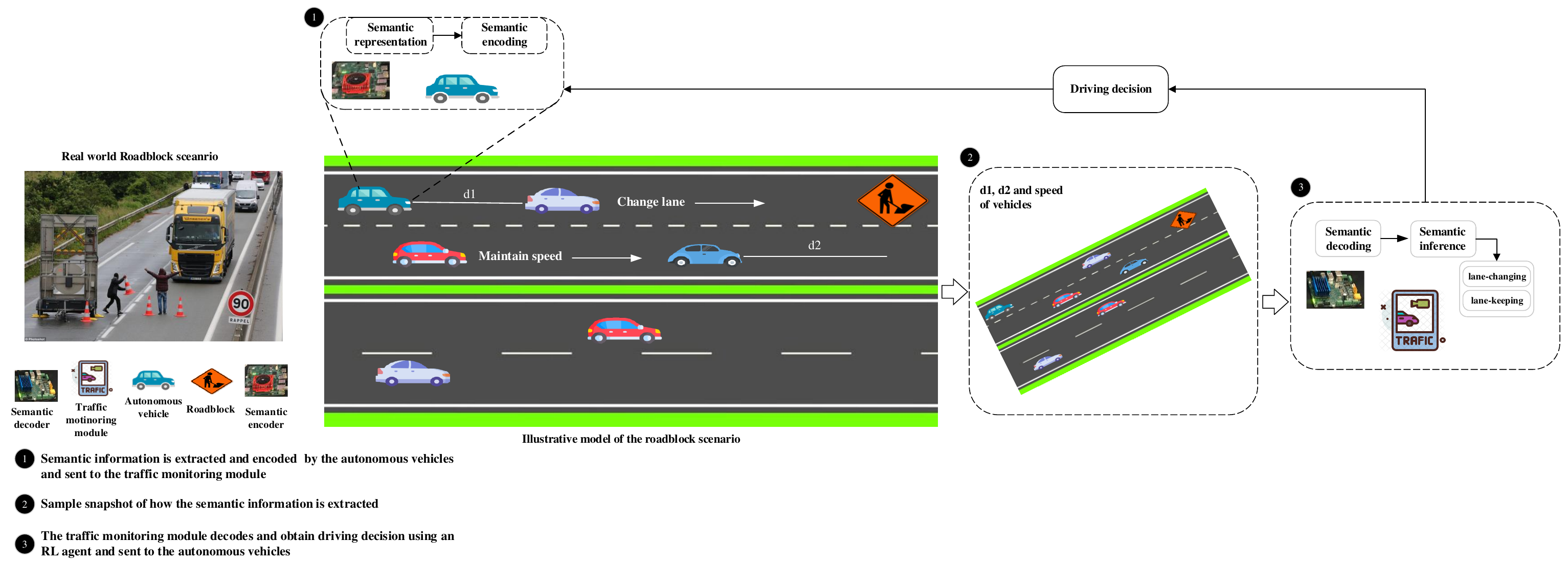}
    \caption{DL-based semantic traffic control}
    \label{fig:Fig2}
\end{figure*}

\section{Background and Related Works}\label{RW}

In this section, we offer a concise overview of the foundational concepts of semantic communication and its application within the Intelligent Transportation Systems (ITS) field, along with a survey of pertinent literature.

\subsection{Semantic Communication}

The emergence of semantic-based communication has introduced novel approaches to address the hurdles encountered by conventional communication systems. Semantic communication possesses the ability to sift through data and extract only the pertinent information within the semantic domain \cite{wang2023road} \cite{yang2022semantic}. This capacity allows it to compress data while preserving its underlying meaning. Furthermore, semantic-based communication is designed to exhibit resilience in noisy environments, rendering it well-suited for a multitude of practical applications across various industries. Consequently, there is a substantial research focus, both in academia and industry, on developing semantic communication systems. In \cite{yang2022semantic}, the authors delineated three categories of semantic communication systems: semantic-oriented communications, goal-oriented communications, and semantic-aware communications. In the context of this study, we specifically address semantic-oriented communications, while readers interested in the other two categories can refer to \cite{yang2022semantic} for more details.

In semantic-oriented communication, the primary emphasis lies on the semantic content of the source data rather than the absolute accuracy of transmitted bits. This approach involves the use of a semantic representation module, tasked with capturing the semantic content from the source data while filtering out any superfluous or redundant information before the standard data encoding process for transmission. Upon reception, semantic inference and semantic decoding are employed to decode the transmitted data, typically with the assistance of potent AI-driven tools such as Neural Networks (NN). This process results in data compression and reduction, thereby enhancing transmission accuracy and efficiency. 


Semantic communication introduces a departure from the conventional physical layered block structures typically found in traditional communication systems. Instead, it embraces end-to-end neural networks, facilitating the realization of joint transceiver optimization, as highlighted in \cite{nan2023physical}. As a result, researchers have advocated for the integration of Deep Neural Networks (DNN) within semantic communication to enhance communication efficiency. More specifically, neural architectures resembling autoencoders \cite{kramer1991nonlinear} play a pivotal role by extracting semantic representations of the data before the process of semantic encoding on the transmitter side. Subsequently, on the receiver side, a semantic encoder is deployed to reconstruct the data. For example, in \cite{xie2021deep}, the authors unveiled a DL-based semantic framework designed for text-based data transmission. This framework is engineered with the objective of optimizing system capacity while concurrently minimizing semantic errors. Additionally, in \cite{zhang2022deep}, the authors present a dynamic DL-based semantic communication system. This system exhibits a dynamic nature, remaining agnostic about the specific task at the transmitter end, particularly in image transmission. In this work, we are particularly interested in using the DL-based semantic framework to extract semantic information that will be passed on the an RL agent to make informed decisions in a driving scenario.




\subsection{Applications of Semantic Communication in ITS domain}


Researchers have recently embraced semantic-based communication frameworks within the ITS domain. For instance, in \cite{raha2023artificial}, the authors introduced an AI-based semantic communication framework tailored for autonomous vehicles. Their approach involves leveraging a traffic infrastructure to monitor the dynamics of vehicles and transmitting this data to a central Base Station (BS). This BS employs a neural network to extract semantic information, which is subsequently employed to convey informed driving decisions to autonomous vehicles, particularly concerning traffic stops along the roadway. In \cite{xu2023semantic}, the authors introduce a collaborative semantic-based communication framework that harnesses cooperative semantic information from vehicles. The primary goal is to reduce the extensive data traffic, particularly within the context of Internet of Vehicles (IoV) in ITS. Another semantic communication framework is presented in \cite{tang2022intelligent}, emphasizing intelligent fabric for autonomous vehicles. This framework enables seamless intelligent interactions within a transportation-in-cabin environment through the deployment of intelligent fabrics.

In contrast to the previously mentioned approaches, our work capitalizes on the semantic-based communication framework to extract semantic information from snapshots of the considered roadblock scenario. We specifically focus on enabling semantic encoding to be executed within the vehicles themselves, eliminating the need to rely on server-based processing and alleviating resource constraints.



\section{System Model}\label{SM}

This section commences with an exposition of the envisioned DL-based semantic traffic control framework. It proceeds to present a comprehensive MDP formulation for the scenario under consideration and culminates with an elucidation of the RL-based solution approach.
\subsection{Proposed Model}

In this subsection, we present the DL-based semantic traffic control system we propose. Figure~\ref{fig:Fig2} illustrates the overarching system model, where we envision an ITS. We address a scenario where a roadblock arises due to factors such as road maintenance or accidents. To accurately depict this scenario, we assume that the vehicles on the road are autonomous, equipped with AI capabilities, and possess wireless communication capabilities like V2V or V2I. Broadly, our system model encompasses autonomous vehicles, and a traffic monitoring module 
as depicted in Figure~\ref{fig:Fig2}.

The AVs have the capability to execute semantic encoding, which allows them to extract the necessary semantic information in the form of observations. The observations are divided into two distinct categories: learnable patterns and memorizable patterns. In accordance with the approach introduced in \cite{chaccour2022less}, the memorizable patterns are characterized by their inherent randomness, necessitating a complex learning process for extracting semantic representations. Consequently, it is advisable to transmit these types of attributes using conventional communication resources. In the context of this study, examples of memorizable patterns include the velocity and location of the vehicles. 
It's essential to note that in Figure~\ref{fig:Fig2}, we exclusively depict the aspect related to semantic communication. The assumption is that memorizable patterns are conveyed using conventional wireless transmission methods. In contrast, the learnable patterns encompass attributes from which semantic information can be extracted. Within this framework, $d_1$ and $d_2$ are notable examples of such patterns.

The traffic monitoring module is endowed with AI capabilities to receive the encoded observations, which encompass traffic scenario attributes such as $d_1$ and $d_2$ representing the distances between two vehicles and the distance between a vehicle and the roadblock, respectively. Additionally, it records the cars' velocities and their positions along the route. The role of the traffic monitoring module extends to decoding the received information and utilizing a DQN agent to obtain appropriate driving decisions (i.e., either lane-changing or lane-keeping) as shown in Figure~\ref{fig:Fig2}. Moreover, these decisions are sent back to the vehicles to act appropriately.

\subsection{MDP Formulation and Solution Approach}
Mathematically, we model the proposed DL-based semantic traffic control as an MDP \cite{x2} as follows.

\begin{enumerate}
    \item States: 
    In the proposed DL-based semantic traffic control model, the state is defined as a tuple $<d_1, d_2, v_i, x_i, y_i>$ where $d_1$ represents the distance between the target vehicle and the vehicle in front of it, $d_2$ represents the distance between the target vehicle and the roadblock, $v_i$ represents the speed of the target vehicle $i$, and $x_i$, $y_i$ represents the location of the target vehicle $i$. It is worth mentioning that a neural network is used to extract this information from the frame of the snapshot.
    \item Actions: 
    The action space is represented as $<A_{lk}, A_{lc}>$, wherein $A_{lk}$ signifies the choice of staying within the same lane, referred to as a "lane-keeping decision," and $A_{lc}$ corresponds to the decision to switch lanes in order to circumvent the roadblock, termed a "lane-changing decision."
    \item Rewards: The definition of rewards is structured as a tuple $<R_1, R_2, R_3>$. Within this framework, $R_1$ signifies a positive reward earned when the target autonomous vehicle successfully switches lanes and seamlessly merges with the average speed before arriving at the roadblock. On the other hand, $R_2$ represents a negative reward, applicable when the target vehicle fails to change lanes after encountering the roadblock, which typically denotes reaching the end of the lane. The third component, $R_3$, embodies a positive reward (which is less than $R_1$), awarded when the target vehicle changes lanes and manages to merge but does so at the minimum speed just before the roadblock. This particular reward mechanism addresses scenarios where the target car ultimately merges, albeit at the very end of the lane, potentially causing substantial traffic congestion.
\end{enumerate}

To solve the MDP model above, we employ a DQN-based algorithm. The pseudocode of the algorithm is given in algorithm 1. In this application, information is derived from the images from the semantic communication modules specifically the encoder and decoder. The encoder extracts the features to then pass them to a decentralized decoder which then is able to take the useful information and pass it to the DQN.

\begin{algorithm}
\caption{Semantic-based DQN algorithm for obtaining optimal driving decisions}\label{alg:DQN}
\begin{algorithmic}[1]
\Require Initial weight (\textbf{w}), Initial replay buffer $D$, Initial policy $\pi(s)$ as $\epsilon$-greedy policy, initial location of the roadblock from semantic encoder; 
\Require Encoder $\rightarrow$ feature\_extraction 
\Require Decoder $\xrightarrow{feature\_extraction}$ Position(Road\_block)
\Require Initial $s_0 =$ Position(Road\_block)
\For{$t = 0, 1, ... ,$}  
    \State Execute action $a_t = \pi(s_t)$ according to $\epsilon$-greedy policy
    \State Observe state $s_{t+1}$ and reward $r_{t+1}$
    \State Store the transition parameters ($s_t, a_t, r_{t+1}, s_{t+1})$ in $D$
    \State Sample minibatch transitions randomly from $D$ as $(s_t, a_t, r_{t+1}, s_{t+1})$
    \State Calculate the target Q value
      \State Update \textbf{w} using stochastic descent to minimize the loss function 
    \EndFor
    \State Improve policy $\pi(s)$ as $\epsilon$-greedy with the new \textbf{w}
\end{algorithmic}
\end{algorithm}

\section{Experimentation}\label{ER}

The experimentation conducted for this paper primarily focused on introducing a method to lower transmitted data between a sensor gathering information and the hardware running the simulation environment for traffic optimization. The traffic optimization model is pre-trained taking in roadblock location as input, and then focuses on enhancing traffic flow optimization, particularly in scenarios where a lane is obstructed within a 3-lane highway. The RL agent underwent training in a dynamic environment where the obstruction was strategically relocated to various positions along the highway. However, it's important to note that this model is just one element within a larger system. The RL algorithm functions by taking the highway's location as input and generating optimal lane change decisions as output. To facilitate this system's operation, there's a need for a mechanism to both receive and transmit the necessary information efficiently.

This is where the semantic system comes into the system model.Instead of transmitting complete images between the two hardware components, where the image must be classified and then used for the RL algorithm, a method of feature extraction is used. This process allows for the extraction of critical information while discarding the non-essential components, thereby reducing the size of the data transmission.

For this experimentation, two Systems on Module(SOM) boards (AMD KV260 and AMD KD240) were used as an encoder and decoder. Both are powerful SOMs however, still have limited resources while utilizing ARM processors and program acceleration. 

\subsection{Encoder and Decoder}

To build an efficient system, a model must be trained to be able to extract information from an image and store it in a smaller file. This part of the system is classified as the encoder, for the system to work a decoder must be paired with it. The decoder will be able to make sense of the features that have been encoded and sent. By using this method, we can reaffirm that the features will not be misinterpreted and differ from the class of the original image. 


\subsection{Training the Object Detection Model}

A custom framework was created using only numpy as an external library, this was done so that only the necessary components of the program would be included greatly reducing bloat from importing external libraries. It also guarantees compatibility between the model and the SOMs.

After appropriate datasets were found, a simple object detection model needed to be trained. Many large libraries such as Tensorflow, Keras, and PyTorch have added support for arm processors, but still have small bugs with dependencies libraries. This requires workarounds for many tasks, to avoid this a minimal ARM framework was created using only numpy to create a small deep-learning model. The framework included only the necessary functions for creation and training, such as initialization of weights given a model size, forward propagation, backward propagation, predictions, time metrics, and saving weights. This provided more stable training preventing, memory overflows and crashes. 

The framework also included a data preprocessing system to optimize storage as well as normalization. To normalize the dataset a uniform size was selected of 32 pixels by 32 pixels and grayscale was applied, and this was done across all three datasets. This was impactful to the model by not only determining a standard input layer between the datasets but also allowing the dataset to be encoded in a single large vector. This helped when separating the label from the data as the first 1024 elements would be the image while the 1025th element would be the label. The following layer is of size 10 along with a set of 10 biases and to avoid local minimums weights are initialized randomly and can be re-initialized. During forward propagation from the first layer to the second layer, the initial random weights and the input layer have the dot product calculated and then the bias is summed to each of the 10 nodes and then passed to the Relu activation function, where it will act as input for the next layer. From the second to the third layer, the process is the same however, the activation function is a soft max which takes the highest value after it has been normalized across all inputs. Backpropagation is what allows learning by correcting the weights at the given nodes by reversing the forward propagation in reverse order transposing the dot products, and taking the derivates of the activation functions. The results are the adjusted weights, which are calculated by taking the current weight and subtracting the backpropagated weight, and multiplying them by the alpha which is a user-defined learning hyperparameter. 

The rest of the additions to the framework can be considered low-complexity additives for ease of use and quality of experience. Three different models were created for the three different datasets used, the KUL Belgium Traffic Sign Classification dataset, the MASTIF Croatian dataset, and the German Traffic Sign Recognition Benchmark. Each of the datasets had different classes which meant that the last layer had to be modified to match the classifiers. Once the model was completed training, the weights were saved using numpy compression file, this would allow for the weights to be loaded in rather than retraining. 

Now that a single model had been trained it was time to split it into an encoder and decoder. The model would be implemented on both devices with the weights from training, however, both would have different entry points in the program. For the encoder, the image would pass through the input layer and the features would be extracted instead of passing the features to the next layer for classification, the output from the model is saved as a numpy compressed file; this file ends up being smaller than the actual image. For the decoder, it receives the compressed feature file loads it in, and then applies the values to the next layer and makes a classification. The decoder also has the weights and model for the traffic optimization on board, once the classification it can pass the information to the traffic DRL model to predict the inputs. By splitting the entry points of the model between two devices a semantic communication system can be achieved while also lowering the file size and transmission time.

\subsection{Network Testbed}

To ensure that the system is operational and would work in implementation, both the SOMs are set up in a Local Area Network, and both firewalls are enabled to allow SSH which will also support the transfer of files. Scripts would be run to collect the time metrics of things such as the run-time of the program and the run-time of the transmission.

\begin{figure}[htb]
    \centering
   \includegraphics[width=.5\textwidth]{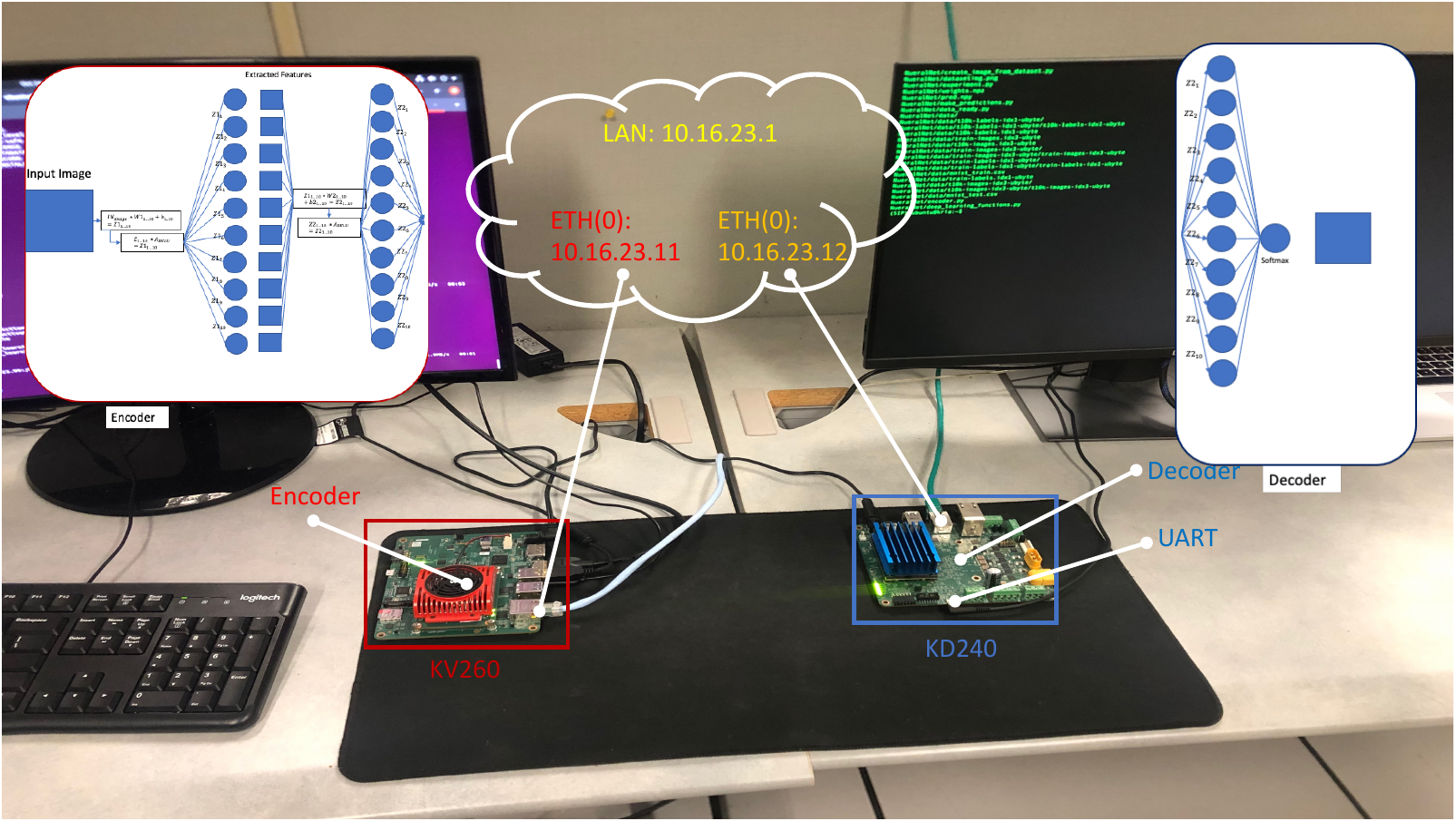}
    \caption{Experiment Setup with KV260 (left) and KD240 (right) AMD Boards}
    \label{fig:Fig3}
\end{figure}

Shown in Figure \ref{fig:Fig3}, is the setup for this experiment with both SOMs. The KV260 is equipped with a graphics driver through an HDMI port so the SOM could be used like a native desktop, however, the KD240 SOM does not have an integrated graphics driver; Therefore serial communication had to be established using the SOM's Universal Asynchronous Receiver and Transmitter (UART) interface. Both the SOMs run a modified version of Ubuntu 22.04 provided by AMD under the name Kria. The figure also gives a visual representation to the entry points of each SOM on the model.

\section{Analysis and Results}

To establish a baseline to compare to, the model would only be run on the decoder where it would make the classification from the image, the encoder would act only as an information relay forwarding the full-size images to the decoder. A single image was sent, 900 images were sent to mimic a 30-second 30 frame per second video, and 1800 images were sent to mimic a 60-second video. The metrics that were collected were the run time to complete the classification through the entire model, the time it took to send the files and the size of the files. Once the benchmark was established the same conditions were applied to the semantic system.

\begin{figure}[htb]
    \centering
   \includegraphics[width=.5\textwidth]{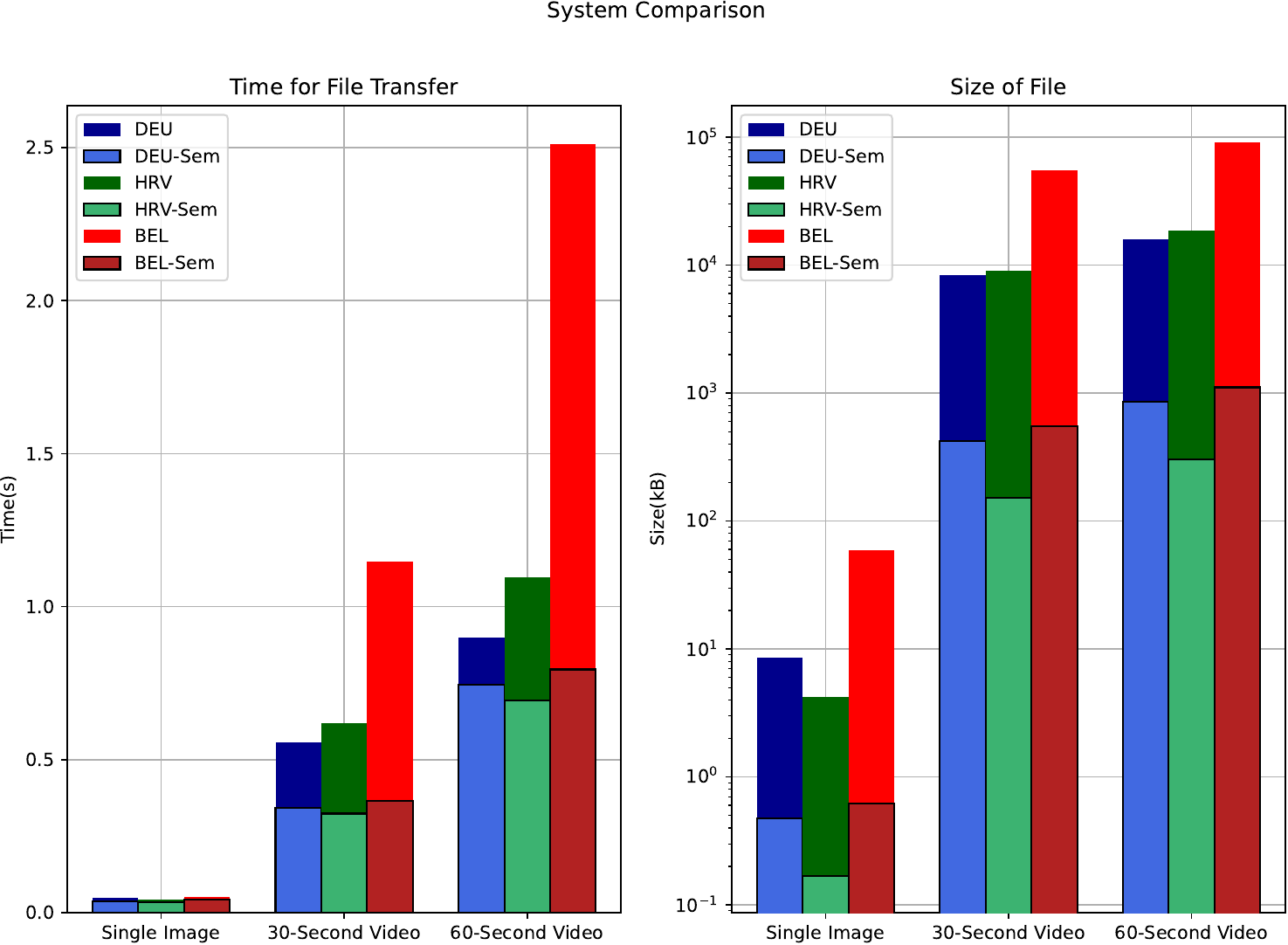}
    \caption{File Size and Time Comparison between systems of the datasets: 
    \centering DEU-German HRV-Croatia BEL-Belgium}
    \label{fig:Fig4}
\end{figure}

Figure \ref{fig:Fig4} highlights the reduction of data between the sending of images normally over a network against the semantic system. The three different models are included in the three scenarios which include sending a single image, 900 images equivalent to a 30-second video, and 1800 images equivalent to a 60-second video. The left side of the figure shows the timing of the file transfer, which shows that sending the images normally takes more time than the systematic system. Variations in the times for the normal system are also due to varying image sizes, however, for the semantic system, all inputs to all models were normalized. The right side of the figure shows us the reason for the reduction in the transfer times is that the files are significantly smaller. 



The SOM that hosted the decoder also had the pre-trained RL algorithm, this allows for the decoded information to be placed as the input for the RL optimization algorithm within the same script. This is more demanding on the SOM but will allow for faster response from the entire system. The model was first tested to ensure its ability to run on the SOM and also test the accuracy. Since the system is operating under the assumption that inputs could be a 30-frame-per-second video, 20 seconds is tested, meaning a total of 900 predictions for optimizing the traffic. 

\begin{figure}[htb]
    \centering
   \includegraphics[width=.5\textwidth]{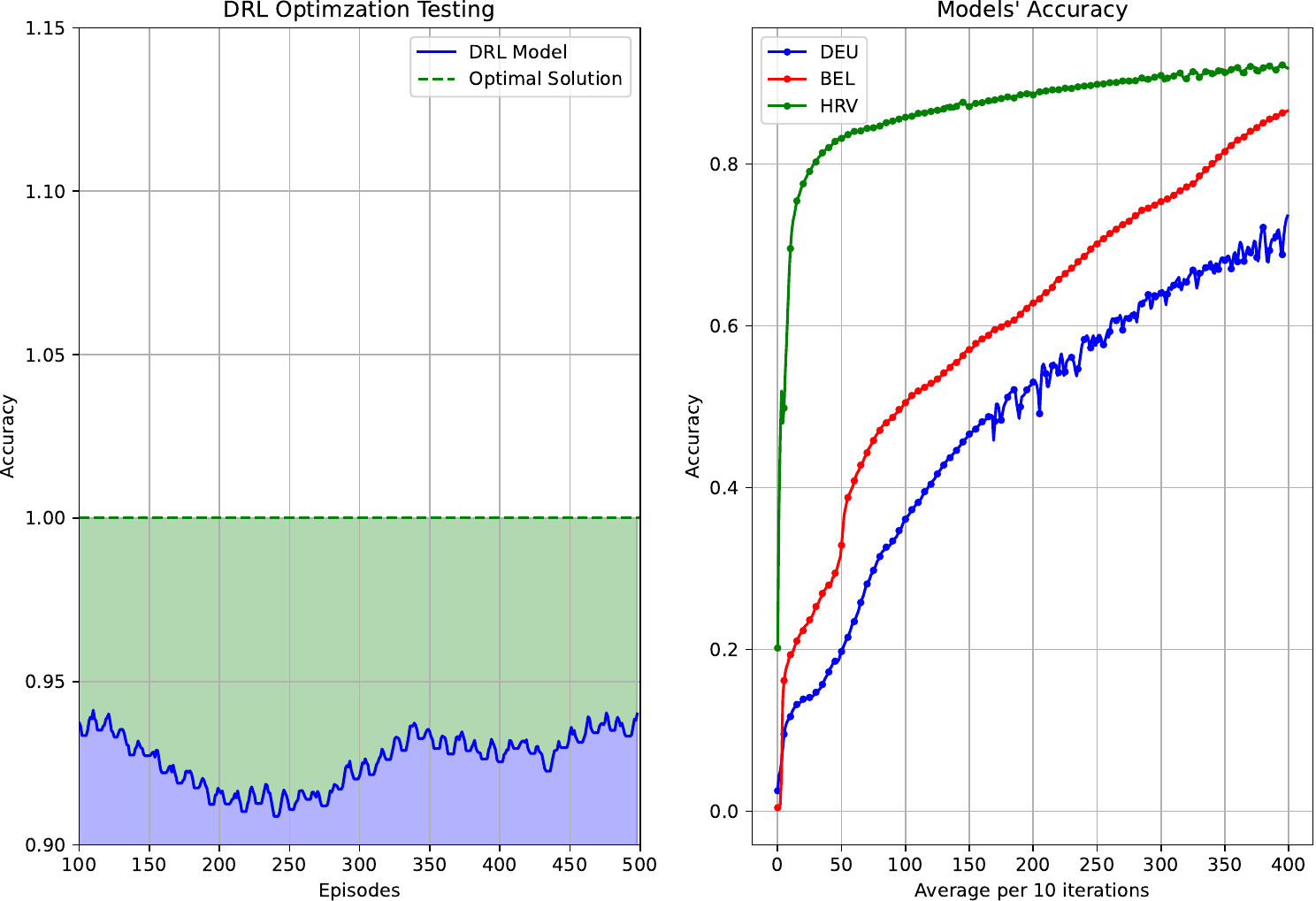}
    \caption{Accuracy of DRL Traffic and Model Accuracy on the three Datasets}
    \label{fig:Fig5}
\end{figure}

The model was programmed under the assumption that the car's max speed is 5 units/s, this value is variable based on the speed limit if the average flow of traffic falls between a certain threshold set by the model, ( 1/5 of the speed limit), a negative reward will be issued. This established the range for the optimal solution, and therefore accuracy could be calculated. The result shown in Figure \ref{fig:Fig5} shows that the optimization model reaches around a 92\% accuracy, this variation is due to noise added during the creation of the simulation, in the form of varying the speed of the vehicles part of the simulation. The varying accuracy in the models is due to the datasets varying size of input data and classes which must be learned.

\section{Conclusion and Future Directions}\label{C}

This work presented a DL-based semantic traffic control system that utilizes semantic encoding to gather, reduce, and improve communication of the system. The semantic system reduces the amount of data needed to be transmitted through feature extraction encoding and decoding and decreases transmission time. This was shown in the experimentation for not only single images but videos as well, the system is also scalable to IoT devices such as SOMs that utilize ARM processors. Future research directions involve gathering different forms of information other than images, such as readings from sensors. This can be accomplished by an ensemble of deep models to filter and extract information from more than one source while keeping the same framework and concepts present in this paper.



\bibliographystyle{IEEEtran}
\bibliography{references}

\begin{thebibliography}{10}
\providecommand{\url}[1]{#1}
\csname url@samestyle\endcsname
\providecommand{\newblock}{\relax}
\providecommand{\bibinfo}[2]{#2}
\providecommand{\BIBentrySTDinterwordspacing}{\spaceskip=0pt\relax}
\providecommand{\BIBentryALTinterwordstretchfactor}{4}
\providecommand{\BIBentryALTinterwordspacing}{\spaceskip=\fontdimen2\font plus
\BIBentryALTinterwordstretchfactor\fontdimen3\font minus
  \fontdimen4\font\relax}
\providecommand{\BIBforeignlanguage}[2]{{%
\expandafter\ifx\csname l@#1\endcsname\relax
\typeout{** WARNING: IEEEtran.bst: No hyphenation pattern has been}%
\typeout{** loaded for the language `#1'. Using the pattern for}%
\typeout{** the default language instead.}%
\else
\language=\csname l@#1\endcsname
\fi
#2}}
\providecommand{\BIBdecl}{\relax}
\BIBdecl

\bibitem{li2023automated}
Q.~Li, X.~Li, H.~Yao, Z.~Liang, and W.~Xie, ``Automated vehicle identification
  based on car-following data with machine learning,'' \emph{IEEE Transactions
  on Intelligent Transportation Systems}, 2023.

\bibitem{liu2023real}
C.~Liu, H.~Yang, M.~Zhu, F.~Wang, T.~Vaa, and Y.~Wang, ``Real-time multi-task
  environmental perception system for traffic safety empowered by edge
  artificial intelligence,'' \emph{IEEE Transactions on Intelligent
  Transportation Systems}, 2023.

\bibitem{luo2022semantic}
X.~Luo, H.-H. Chen, and Q.~Guo, ``Semantic communications: Overview, open
  issues, and future research directions,'' \emph{IEEE Wireless
  Communications}, vol.~29, no.~1, pp. 210--219, 2022.

\bibitem{xie2021deep}
H.~Xie, Z.~Qin, G.~Y. Li, and B.-H. Juang, ``Deep learning enabled semantic
  communication systems,'' \emph{IEEE Transactions on Signal Processing},
  vol.~69, pp. 2663--2675, 2021.

\bibitem{huang2021deep}
D.~Huang, X.~Tao, F.~Gao, and J.~Lu, ``Deep learning-based image semantic
  coding for semantic communications,'' in \emph{2021 IEEE Global
  Communications Conference (GLOBECOM)}.\hskip 1em plus 0.5em minus 0.4em\relax
  IEEE, 2021, pp. 1--6.

\bibitem{figetakis2023implicit}
E.~Figetakis, Y.~Bello, A.~Refaey, L.~Lei, and M.~Moussa, ``Implicit sensing in
  traffic optimization: Advanced deep reinforcement learning techniques,''
  \emph{arXiv preprint arXiv:2309.14395}, 2023.

\bibitem{wang2021harmonious}
G.~Wang, J.~Hu, Z.~Li, and L.~Li, ``Harmonious lane changing via deep
  reinforcement learning,'' \emph{IEEE Transactions on Intelligent
  Transportation Systems}, vol.~23, no.~5, pp. 4642--4650, 2021.

\bibitem{wang2023road}
C.-X. Wang, X.~You, X.~Gao, X.~Zhu, Z.~Li, C.~Zhang, H.~Wang, Y.~Huang,
  Y.~Chen, H.~Haas \emph{et~al.}, ``On the road to 6g: Visions, requirements,
  key technologies and testbeds,'' \emph{IEEE Communications Surveys \&
  Tutorials}, 2023.

\bibitem{yang2022semantic}
W.~Yang, H.~Du, Z.~Q. Liew, W.~Y.~B. Lim, Z.~Xiong, D.~Niyato, X.~Chi, X.~S.
  Shen, and C.~Miao, ``Semantic communications for future internet:
  Fundamentals, applications, and challenges,'' \emph{IEEE Communications
  Surveys \& Tutorials}, 2022.

\bibitem{nan2023physical}
G.~Nan, Z.~Li, J.~Zhai, Q.~Cui, G.~Chen, X.~Du, X.~Zhang, X.~Tao, Z.~Han, and
  T.~Q. Quek, ``Physical-layer adversarial robustness for deep learning-based
  semantic communications,'' \emph{IEEE Journal on Selected Areas in
  Communications}, 2023.

\bibitem{kramer1991nonlinear}
M.~A. Kramer, ``Nonlinear principal component analysis using autoassociative
  neural networks,'' \emph{AIChE journal}, vol.~37, no.~2, pp. 233--243, 1991.

\bibitem{zhang2022deep}
H.~Zhang, S.~Shao, M.~Tao, X.~Bi, and K.~B. Letaief, ``Deep learning-enabled
  semantic communication systems with task-unaware transmitter and dynamic
  data,'' \emph{IEEE Journal on Selected Areas in Communications}, vol.~41,
  no.~1, pp. 170--185, 2022.

\bibitem{raha2023artificial}
A.~D. Raha, M.~S. Munir, A.~Adhikary, Y.~Qiao, S.-B. Park, and C.~S. Hong, ``An
  artificial intelligent-driven semantic communication framework for connected
  autonomous vehicular network,'' in \emph{2023 International Conference on
  Information Networking (ICOIN)}.\hskip 1em plus 0.5em minus 0.4em\relax IEEE,
  2023, pp. 352--357.

\bibitem{xu2023semantic}
W.~Xu, Y.~Zhang, F.~Wang, Z.~Qin, C.~Liu, and P.~Zhang, ``Semantic
  communication for the internet of vehicles: A multiuser cooperative
  approach,'' \emph{IEEE Vehicular Technology Magazine}, vol.~18, no.~1, pp.
  100--109, 2023.

\bibitem{tang2022intelligent}
Y.~Tang, N.~Zhou, Q.~Yu, D.~Wu, C.~Hou, G.~Tao, and M.~Chen, ``Intelligent
  fabric enabled 6g semantic communication system for in-cabin scenarios,''
  \emph{IEEE Transactions on Intelligent Transportation Systems}, vol.~24,
  no.~1, pp. 1153--1162, 2022.

\bibitem{chaccour2022less}
C.~Chaccour, W.~Saad, M.~Debbah, Z.~Han, and H.~V. Poor, ``Less data, more
  knowledge: Building next generation semantic communication networks,''
  \emph{arXiv preprint arXiv:2211.14343}, 2022.

\bibitem{x2}
E.~Figetakis, A.~R. Hussein, and M.~Ulema, ``Evolved prevention strategies for
  6g networks through stochastic games and reinforcement learning,'' \emph{IEEE
  Networking Letters}, vol.~5, no.~3, pp. 164--168, 2023.

\end{thebibliography}


\end{document}